\documentclass[reprint, cha]{revtex4-1}

\usepackage{amssymb}
\usepackage[english]{babel}    % para texto em Inglês
\usepackage[dvips]{graphics}
\usepackage{subfigure}
\usepackage[pdftex]{graphicx}
\usepackage{multirow}
\usepackage{rotating}
\usepackage{picinpar}
\usepackage{color}
\usepackage{lineno}
\usepackage{listings}
\usepackage{epsfig}

\begin{document}

\title{Texture analysis by multi-resolution fractal descriptors}

\author{Jo\~{a}o B. Florindo}
 	     \email{jbflorindo@ursa.ifsc.usp.br}
\affiliation{Scientific Computing Group, S\~ao Carlos Institute of Physics, University of S\~{a}o Paulo (USP),  cx 369 13560-970 S\~{a}o Carlos, S\~{a}o Paulo, Brazil - www.scg.ifsc.usp.br}

\author{Odemir M. Bruno}
              \email{bruno@ifsc.usp.br}
\affiliation{Scientific Computing Group, S\~ao Carlos Institute of Physics, University of S\~{a}o Paulo (USP),  cx 369 13560-970 S\~{a}o Carlos, S\~{a}o Paulo, Brazil - www.scg.ifsc.usp.br}

\date{\today}

\begin{abstract}
This work proposes a texture descriptor based on fractal theory. The method is based on the Bouligand-Minkowski descriptors. We decompose the original image recursively into 4 equal parts. In each recursion step, we estimate the average and the deviation of the Bouligand-Minkowski descriptors computed over each part. Thus, we extract entropy features from both average and deviation. The proposed descriptors are provided by the concatenation of such measures. The method is tested in a classification experiment under well known datasets, that is, Brodatz and Vistex. The results demonstrate that the proposed technique achieves better results than classical and state-of-the-art texture descriptors, such as Gabor-wavelets and co-occurrence matrix.
\end{abstract}


\maketitle

%% main text
\section{Introduction}

Fractal theory plays a fundamental role as an auxiliary tool in the solution of problems in areas as different as Medicine \cite{TWZ07,LC10,LSSMDB10}, Physics \cite{SMS10,HWZ08,CCWH10}, Engineering \cite{CDHLAB03,W08,DAGGR09}, among many others. Particularly, in tasks involving texture analysis, fractal geometry is a powerful modelling tool, achieving interesting results in the description and discrimination of such textures.

In the last two decades, some different fractal approaches to deal with texture analysis have arisen, for instance, multifractals \cite{H01,LRAJ08,LGS00}, the multiscale fractal dimension \cite{MCSM02,CC00}, the fractal descriptors \cite{BPFC08,BCB09,PPFVOB05,FCB10}, among others. Here, we are interested in the fractal descriptors approach.

The main idea of fractal descriptors is to extract a set of features from the estimation of fractal dimension under different scales. Generally, the fractal dimension is based on a power-law relation which expresses the fractality of a structure as a function of measure scale. Unlike the fractal dimension which is a single value, the fractal descriptors are computed over the whole power-law curve \cite{FlorindoBCB12, FlorindoB11}.

An example that illustrates the power of fractal descriptors is showed in \cite{BCB09}. In that solution, the values in the power-law of Bouligand-Minwkowski fractal dimension are used to compose a feature vector to discriminate among plant leaf textures. Actually, this method demonstrates to be successful in the discrimination of natural textures. Such kind of texture present an intrinsic self-similarity property which is notedly well represented by fractal modeling.

Despite their good results, conventional Bouligand-Minkowski fractal descriptors present still a limitation in the representation of textures, mainly when these textures present a higher degree of complexity. This limitation is due mainly to the fact that the descriptors are obtained from the global image, without a more specific treatment of local characteristics present in any real image. Thus, we can obtain more information by estimating those descriptors in different scales over the image.

Considering this assumption, the present work proposes a solution to extract fractal descriptors from a texture based on Bouligand-Minkowski method. Here, we propose the estimation of Bouligand-Minkowski descriptors at different scales (decomposition levels) of the image. The idea is to decompose recursively the image into 4 equal parts and, in each recursion step, we calculate an average and a deviation of the Bouligand-Minkowski descriptors. Thus, from both average and deviation descriptors, we extract entropy measures and compose the feature vector for the texture image.

The method is tested over well-known benchmark texture datasets in a classification task and the results are compared to classical and state-of-the-art texture features methods in the literature, like Gabor wavelets \cite{MM96}, Laws energy \cite{L84}, Gray Level Difference Matrix \cite{WDR76}, etc. The results confirmed the better accuracy of the proposed technique and pointed to the possibility of using the proposed method in a large number of problems involving the description and/or discrimination of textures.

This work id divided into 7 sections, including this introduction. The following provides mathematical background of fractal theory. The third section shows the original Bouligand-Minkowski fractal descriptors. The fourth presents the proposed method. The following explains the experiments. The sixth shows the results of experiments and the final section does the conclusions.

\section{Fractal Geometry}

The literature shows some works applying fractal geometry to solve problems related to texture analysis in a large number of applications \cite{TWZ07,SMS10,CDHLAB03}. The importance of fractals in such kind of tasks is explained by the most flexible representation model provided by fractal theory. In this way, the fractal representation allows the extraction of measures which may describe more faithfully the original structure depicted in the texture image.

Most of such measures is based on the concept of fractal dimension (FD). The importance of FD is due to the fact that it captures the complexity of a fractal object or still, its spatial occupation. Furthermore, these properties are also related to the visual aspect of a texture. Thus, fractal geometry enables a link between the mathematical relations inside a pixel structure and the subjective concept of visual distinction. This link turns fractals into a particularly interesting tool for texture representation and description.

The original definition of fractal dimension is also known as Hausdorff-Besicovitch dimension $dim_{H}$. It is calculated for a set $X\in\Re^{n}$ by
\begin{equation}
dim_{H}(X)=\inf\left\{ s:H^{s}(X)=0\right\} =\sup\left\{ H^{s}(X)=\infty\right\} ,
\end{equation}
in which $H^{s}(X)$ is the $s$-dimensional Hausdorff measure, defined through: 
\begin{equation}
H^{s}(X)=\lim_{\delta\rightarrow0}\inf\left\{ \sum_{i=1}^{\infty}{|U_{i}|^{s}:{U_{i}\mbox{ is a \ensuremath{\delta}-cover of X}}}\right\},
\end{equation}
where $\| \|$ corresponds to the diameter in $\Re^{n}$, that is, $|U|=sup{|x-y|:x,y\in U}$.

In most practical applications, we are interested in calculating the fractal dimension of objects which are not exactly fractals. This is the case here where we are considering texture surfaces, which may be only approximated to fractal objects, if we weaken the infinite self-similarity criterion. A serious drawback in the above fractal dimension definition is that it cannot be applied to these cases where we have not a real fractal. To solve this issue, the literature presents a lot of estimation methods, which compute a fractal dimension value for the real world object, approximating the original fractal dimension concept. Most of such methods is based on the general expression:
\begin{equation}\label{FD}
D(X)=\lim_{\epsilon\rightarrow0}\frac{log(N(\epsilon))}{log(\frac{1}{\epsilon})},
\end{equation}
where $N(\epsilon)$ is a specific measure (depending on the estimation method) of the object and $\epsilon$ is a scale parameter under which the measure is taken \cite{F86}.

We may find a large number of fractal dimension estimation methods \cite{F86}, like box-couting, Bouligand-Minkowski, Fourier, etc. The present work is focused on Bouligand-Minkowski method.

\section{Bouligand-Minkowski Fractal Descriptors}

The Bouligand-Minkowski fractal dimension is obtained by replacing $N(\epsilon)$ in $\ref{FD}$ by a dilation volume $V(r)$. So, in this approach, initially, we map the grayscale image $Img \in [1:M] \times [1:N] \rightarrow \Re$ onto a 3D surface:
\begin{equation}
	Surf = \{i,j,f(i,j)|(i,j) \in [1:M] \times [1:N]\},
\end{equation}
where:
\begin{equation}
	f(i,j) = \{1,2,...,max\_gray\}|f = Img(i,j),
\end{equation}
in which $max\_gray$ is the maximum pixel intensity.

\begin{figure}[!htbp] % Figuras lado a lado
					 \centering
           \epsfig{figure=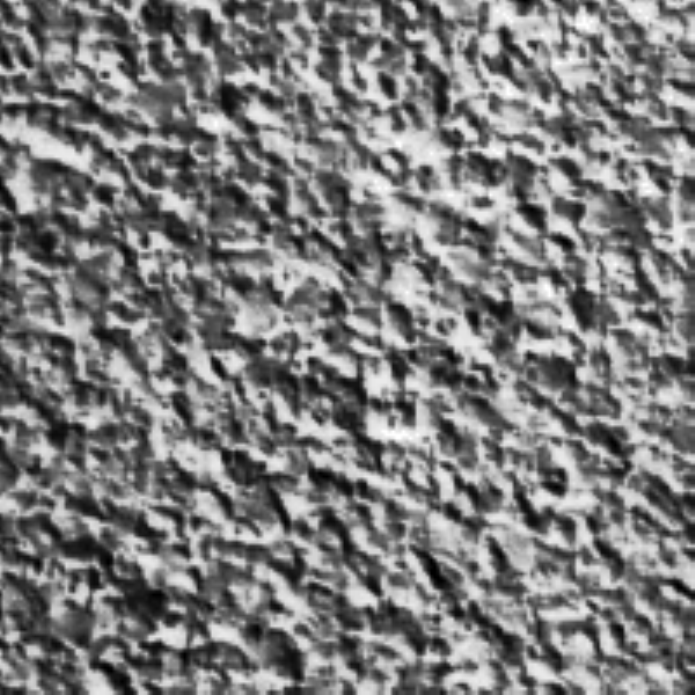,width=0.4\columnwidth}(a)
                 \epsfig{figure=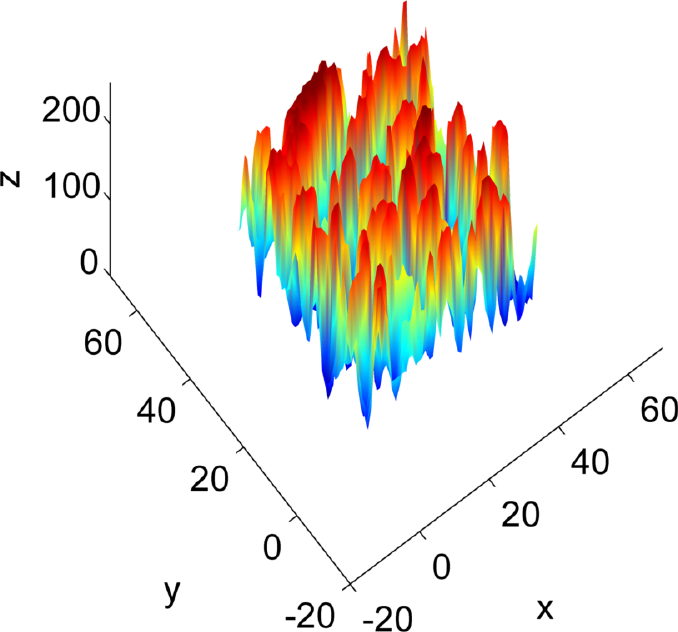,width=0.6\columnwidth}(b)
                 \epsfig{figure=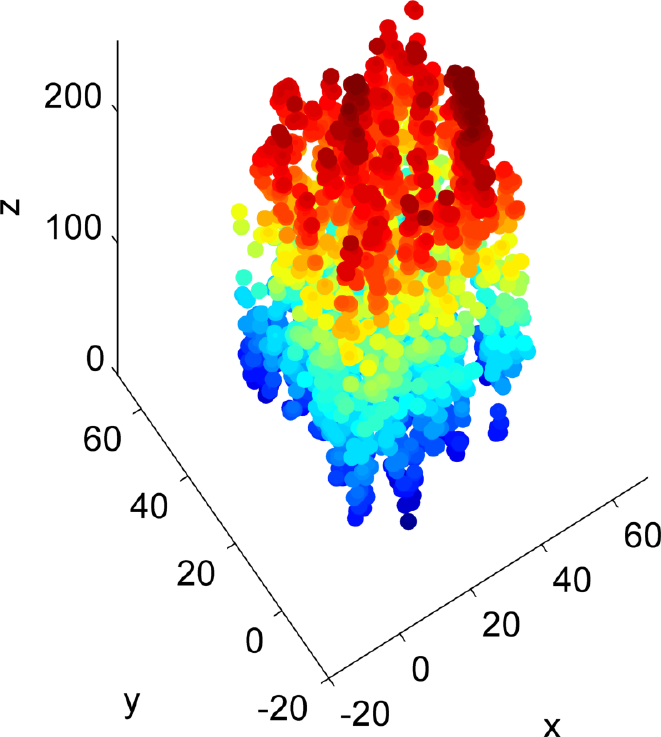,width=0.6\columnwidth}(c)       			 
           \caption{Bouligand-Minkowski fractal dimension estimation. a) Original texture. b) The gray-level image is mapped onto a surface in x-y-z coordinates. c) Each point in the surface is dilated through a sphere with radius $r$.}
           \label{fig:mink}                                  
   \end{figure}

Thus, the mapped surface is summited to a dilation process, by a dilation radius $r$. Essentially, this operation consists in draw spheres with radius $r$ and with center in each point of $Surf$. The dilation volume $V(r)$ corresponds to the total amount of points pertaining to the union of the spheres. The Bouligand-Minkowski dimension is calculated from the curve $\log(V(r)) \times \log(r)$. The Figure \ref{fig:mink} illustrates the process.

Another possible interpretation for the dilation volume is that it corresponds to the number of points with a distance at most $r$ from the original object, in our case, the surface. In this way, the exact Euclidean distance transform (EDT) \cite{FCTB08} is used to optimize the computation of $V(r)$.

The EDT is a transform which maps each point in the 3D space to the distance of this point to a subset of the space. This subspace corresponds to the object of interest, defined by the user. Here, this subset is the mapped surface and the EDT for each point outside $Surf$ is defined through:
\begin{equation}
	EDT(p) = \min\{d(p,q)|q \in Surf^{c}\},
\end{equation}
where $d$ represents the euclidean distance.

In digital images like here, we use the exact EDT and the distances present discrete values $E$:
\begin{equation}
	E = {0,1,\sqrt{2},...,l,...},
\end{equation}
where
\begin{equation}
	l \in D = \{d | d = (i^2+j^2)^{1/2};i,j \in \mathbb{N}\}
\end{equation}

The dilation volume is obtained through:
\begin{equation}
	V(r) = \sum_{i=1}^{r}{Q(i)},
\end{equation}
where
\begin{equation}
	Q(r) = {(x,y,z)|g_k(P) - [g_r(P) \cap \cup_{i=0}^{r-1}g_i(P)]},
\end{equation}
where:
\begin{equation}
	 g_r(P) = 
	  \begin{array}{l} 
	 (x,y,z)|[(x-P_x)^2 + (y-P_y)^2 \\
	 + (z-P_z)^2]^{1/2} = E(r);i,j \in N 
	 \end{array} ,
\end{equation}
where
\begin{equation}
	P = {(x,y,z) | f(x,y,z) \in Surf}
\end{equation}

\begin{figure*}[!htpb]
\centering
\includegraphics[width=\textwidth]{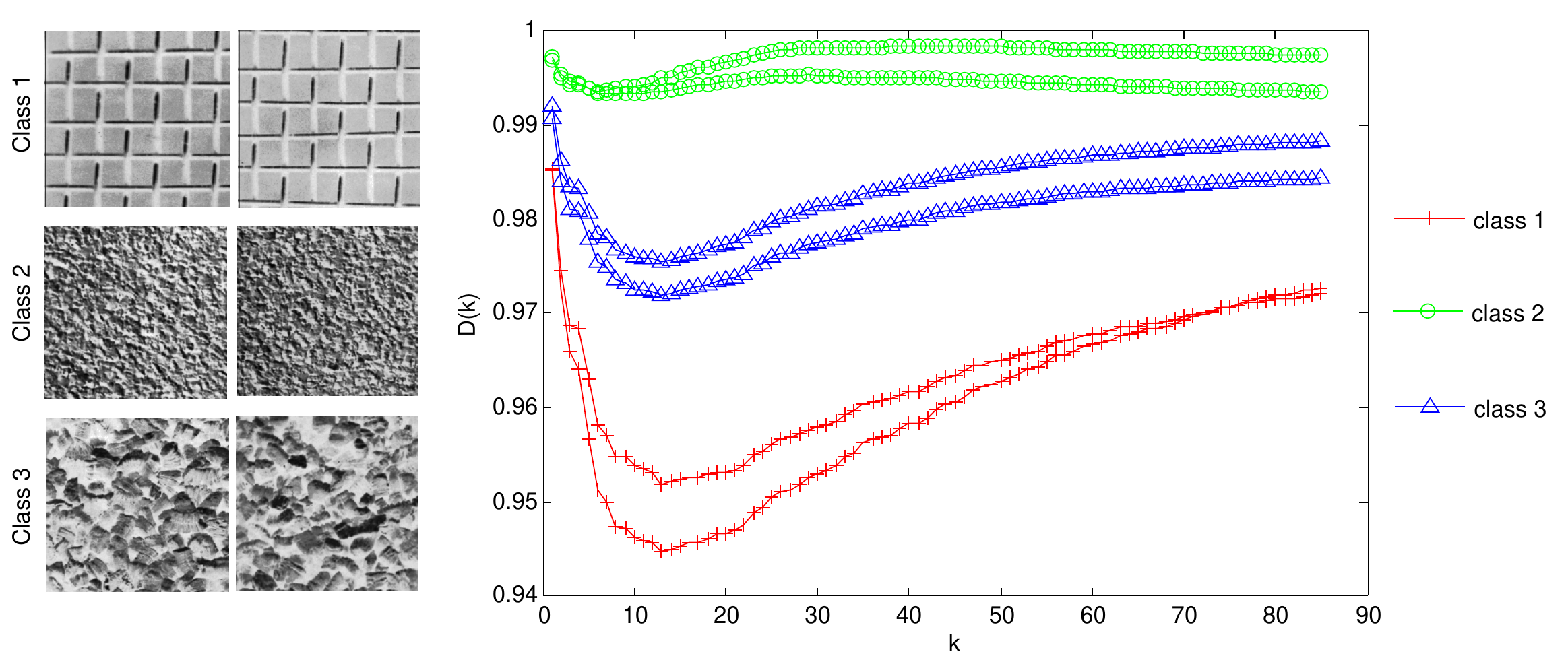}
\caption{Discrimination power of Bouligand-Minkowski fractal descriptors. At left, three classes with two textures in each one. At right, the fractal descriptors curves in a same graph. Observe the visual distinction among classes.}
\label{fig:minkdiscrim}
\end{figure*}

The Bouligand-Minkowski fractal descriptors, defined in \cite{BCB09} correspond to the values of $V(r)$. The Figure \ref{fig:minkdiscrim} shows an example of Bouligand-Minkowski extracted from different textures and the discrimination power of such features.

\section{Proposed Method}

Here, we propose the decomposition of the original texture image into decreasing cell sizes, followed by the calculus of Bouligand-Minkowski descriptors in each cell. The idea is in some way similar to that found in some classical multiscale approaches, like discrete wavelet transform or Gauss pyramid.

The essential idea is to divide recursively the image into 4 equal parts. Each step in this process constitutes a decomposition level. At each decomposition level, we take the average and the standard deviation of descriptors in each cell. Thus, we construct a feature vector from the entropy measure of such descriptors. Finally, we apply a simple attribute selection approach to the feature vector to compose the final descriptors.

Thus, we start with a digital image $I:[N \times N] \rightarrow \Re$. This image is decomposed into levels 
$l|1 \leq l \leq l_{max}$, where $l_{max}$ is the maximum possible level in the image, given by $l_{max} = \mbox{ceil}(\log_2(N))$. In each decomposition level, the image is partitioned into equal regions $R_{ljk}$:
\[
	R_{ljk} = \{x,y|(j-1)*2^{l} \leq x \leq (j)*2^{l},(k-1)*2^{l} \leq x \leq (k)*2^{l}\}.
\]
In each region $R$, we apply the procedure described in the above section and obtain the Bouligand-Minkowski descriptors $D_{ljk}$. For each level $l$, we obtain the average descriptors $D^M_l$ and deviation descriptors $D^{\sigma}_l$:
\[
	D^M_l = \frac{\sum_{jk}{D_{ljk}}}{2^l},
\]
\[
	D^{\sigma}_l = \sum{(D_{ljk}-D_M(l))^2}.
\]

In the following, we extract entropy features from both average and deviation descriptors in each level. Iniatially, for each component (index) $i$ of Bouligand-Minkowski average descriptors at all levels, we construct another vector $\vec{\varphi(i)}$, that is:
\[
	\vec{\varphi(i)} = [D_{1}^{M}(i) D_{2}^{M}(i) D_{3}^{M}(i) ... D_{l_{max}}^{M}(i)].
\]
In the same fashion, we construct the vectors $\vec{\psi(i)}$, from deviation descriptors:
\[
	\vec{\psi(i)} = [D_{1}^{\sigma}(i) D_{2}^{\sigma}(i) D_{3}^{\sigma}(i) ... D_{l_{max}}^{\sigma}(i)].
\]
Then, we compute one Shannon entropy value for each vector. In order to simplify the notation, we call $u$ a generic vector. The entropy is estimated through:
\[
	K(u) = \sum_{i=1}^{N}{u(i)\log(u(i))},
\]
where $N$ is the length of $u$.

The entropy feature vector $\vec{EFV}$ is given by:
\[
	\vec{EFV} = [K(\vec{\varphi(1)}) K(\vec{\varphi(2)}) ... K(\vec{\varphi(n)}) K(\vec{\psi(1)}) K(\vec{\psi(2)}) ... K(\vec{\psi(n)})],
\]
where $n$ is the number of components in average and deviation descriptors.

The final step consists in applying a basic feature selection approach to increase the performance and reduce the number of descriptors. We developed a selection based on the classifier method. Therefore, we computed the classification success rate $S$ for each component of $\vec{EFV}$ vector and sorted the success rates in descending order. We are particularly interested in the vector of sorting indices $\vec{\mathfrak{I}}$:
\[
	\vec{\mathfrak{I}} = \mbox{index}(\mbox{sort}(S(\vec{EFV(i)}))).
\]
Finally, we compose the proposed Multilevel Descriptors $MLD$ by indexing $\vec{EFV}$ through the first $n$ indices in $\vec{\mathfrak{I}}$, where $n$ is the minimum necessary number of components to provide the best possible result. Formally, we have:
\[
	MLD = \mbox{max}_{i}(S(\vec{EFV(\vec{\mathfrak{I}}(1..i)}))).
\] 

\begin{figure}[!htbp]
\centering
\includegraphics[width=\columnwidth]{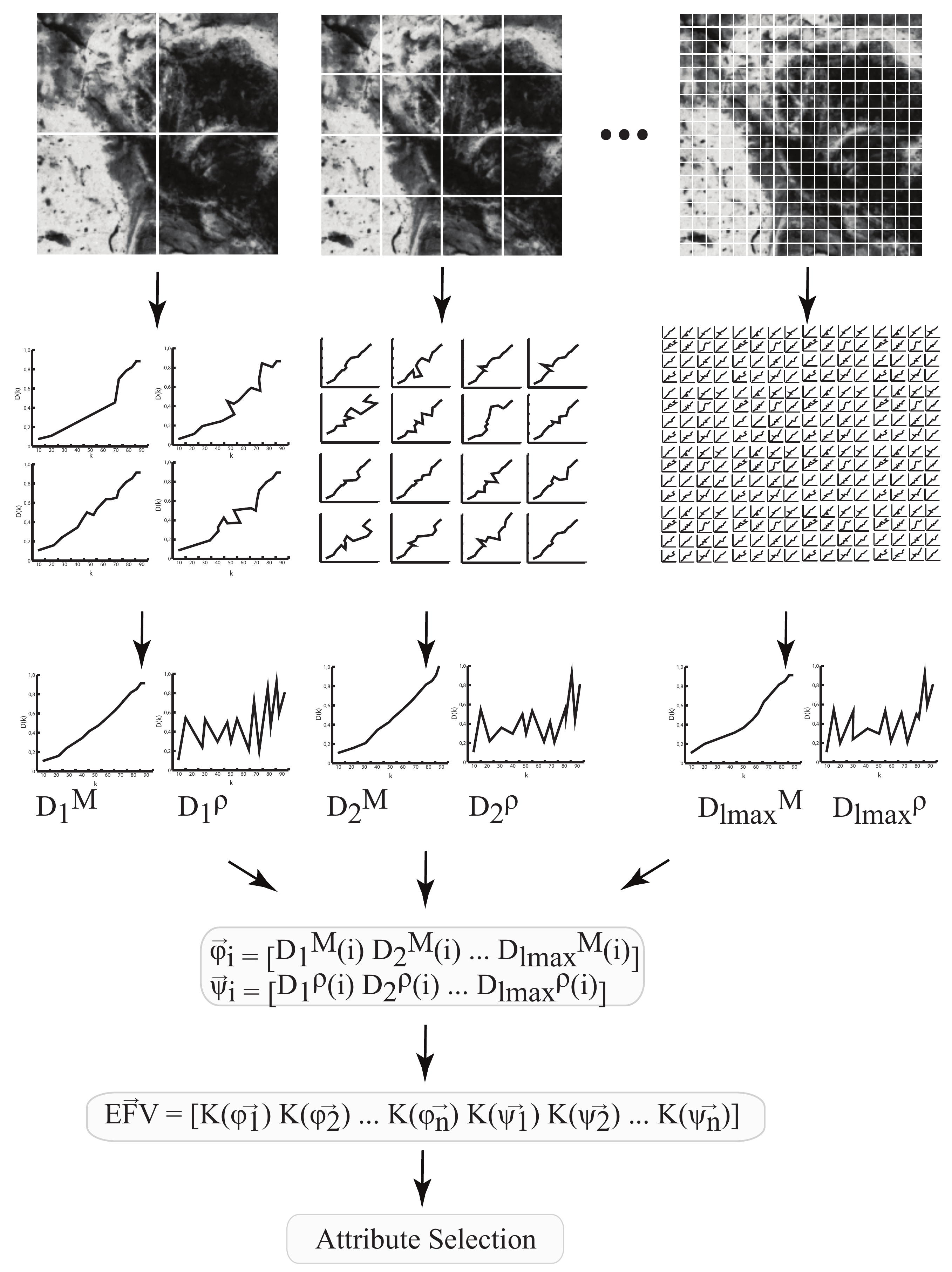}
\caption{Illustration of the composition of feature vector in the proposed technique.}
\label{fig:method}
\end{figure}

\begin{figure*}[!htpb]
\centering
\includegraphics[width=\textwidth]{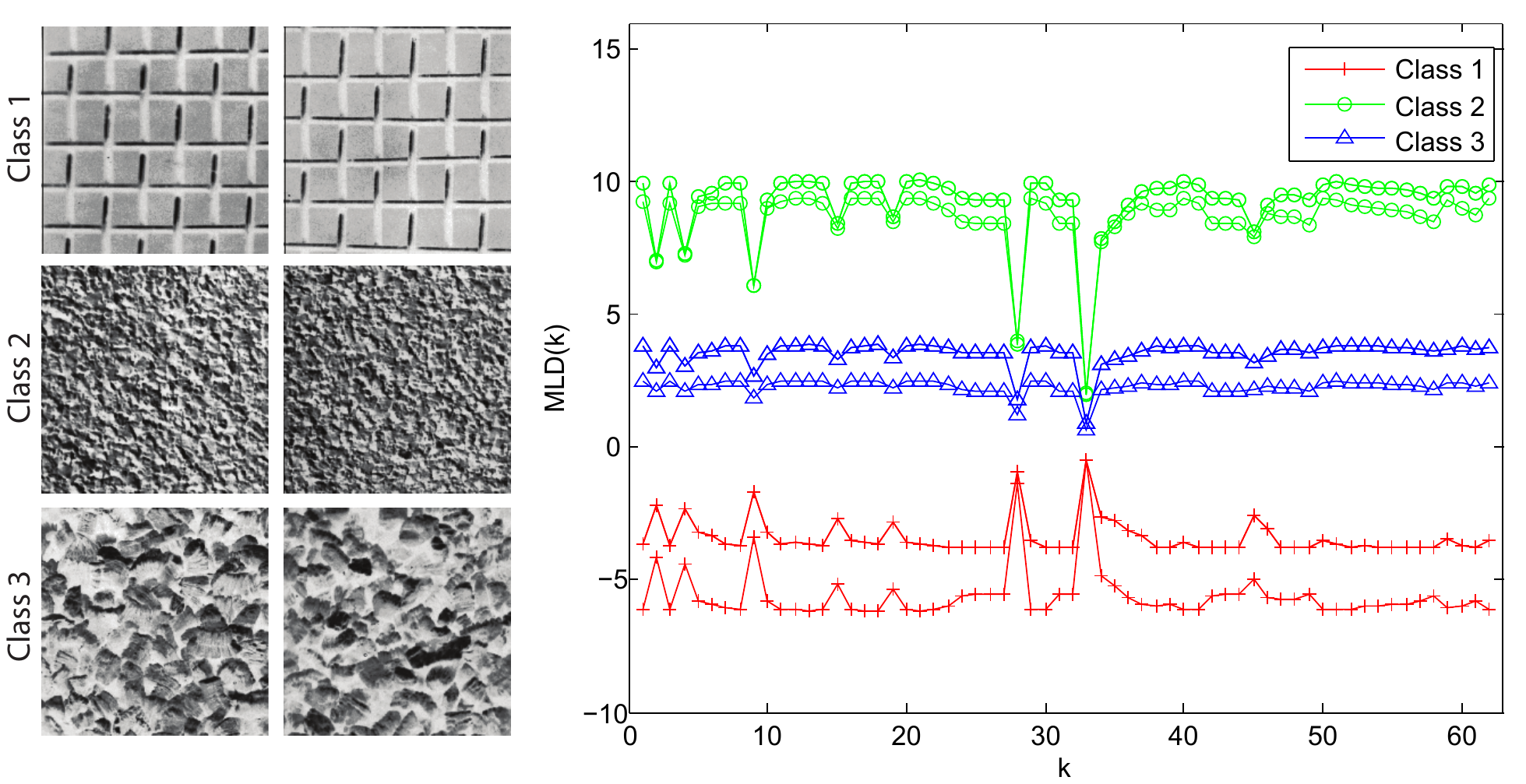}
\caption{Discrimination power of the proposed descriptors. At left, three classes with two textures in each one. At right, the curves of proposed descriptors in a same graph.}
\label{fig:MLDdiscrim}
\end{figure*}

The Figure \ref{fig:method} shows a diagram depicting the whole process.
The Figure \ref{fig:MLDdiscrim} illustrates the power of the proposed method in discriminating among textures from 3 classes in Brodatz benchmark dataset.

\section{Experiments}

In order to verify the efficience of the proposed method, we compared its performance with other texture descriptors in the classification of well known texture data sets.

In the first experiment we used the Brodatz data set \cite{B66}. This is composed by 111 images photographed from an architecture book. In the database, each image is divided into 10 windows 200$\times$200 and the images correspond to the classes. Brodatz data is broadly used as a benchmark set in computer vision and pattern recognition techniques, given the variety of characteristics found in its images, like variations in luminance, geometrical configurations, fidelity to real world textures, among others.

In the following, we classified the Vistex database \cite{Vistex}, a set of natural color textures, composed by 54 classes with 16 samples in each class. Each sample is represented in a 128$\times$128 image. Here, we used the gray-level version of the texture images.

For a fair comparison, we extracted some different classical and state-of-the-art texture descriptors found in the literature and applied to the classifier. The compared methods are Gabor wavelets \cite{MM96}, Co-occurrence matrix \cite{H67}, Laws energy \cite{L84}, Gray Level Difference Matrix (GLDM) \cite{WDR76}, multifractal spectrum \cite{H01} and original Bouligand-Minkowski fractal descriptors \cite{BCB09}. We used the Linear Discriminant Analysis (LDA) classifier \cite{DH00} and adopted the hold-out technique as a statistical training and validating scheme.

\section{Results}

The following Tables \ref{tab:brod} and \ref{tab:vistex} show the results in terms of correctness rate and some other statistical metrics in the classification of the benchmark datasets. We employed the following metrics: Correctness Rate (CR), Kappa index ($\kappa$), Average Correctness Reliability (ACR), Average Error Reliability (AER), Average Error type 1 (AE1) and Average Error type 2 (AE2). The Table \ref{tab:stat} shows a brief description of each metric. More details about each one may be found in \cite{BCB09}. We also show the number of descriptors (ND) employed by each compared approach.
\begin{table*}[!htpb]
	\tiny
	\scriptsize
		\begin{tabular}{|ll|}
			\hline
      	Metric & What measures\\
        \hline
        \hline
        CR & Percentage of elements correctly classified\\
				$\kappa$ & Precision gain relative to a hypothetical random classification\\				
				ACR & Average \emph{a posteriori} probability of correctly classified elements\\				
				AER & Average \emph{a posteriori} probability of misclassified elements\\				
				AE1 & Probability of elements classified as being from any class $j \neq i$ when pertain to the class $i$\\				
				AE2 & Probability of elements classified as being from the class $i$ when pertain to any other class $j \neq i$\\				
			\hline			
		\end{tabular}
	\caption{A brief summary of statistical measures employed in the performance analysis of methods.}
	\label{tab:stat}
\end{table*}
\begin{table*}[!htpb]
	\centering
	\scriptsize
		\begin{tabular}{cccccccc}
			\hline
                 Method & ND & CR (\%) & $\kappa$ & ACR & AER & AE1 & AE2\\
                 \hline
                 \hline
                  Gabor & 20 & 90.09 & 0.90 & 0.98 & 0.84 & 0.08 & 0.10\\
                  Co-occurrence & 84 & 92.07 & 0.92 & 0.99 & 0.91 & 0.07 & 0.08\\
                  GLDM & 20 & 84.14 & 0.84 & 0.94 & 0.79 & 0.15 & 0.16\\
                  Laws & 15 & 87.03 & 0.87 & 0.93 & 0.68 & 0.11 & 0.13\\
                  Multifractal & 101 & 37.48 & 0.37 & 0.92 & 0.86 & 0.63 & 0.62\\
                  Bouligand-Minkowski & 85 & 98.92 & 0.99 & 1.00 & 0.88 & 0.01 & 0.01\\
                  Proposed method & 62 & \underline{99.28} & 0.99 & 1.00 & 1.00 & 0.01 & 0.01\\
			\hline			
		\end{tabular}
	\caption{Correctness rate for Brodatz dataset. The best result is underlined.}
	\label{tab:brod}
\end{table*}

The first point to be observed is that the proposed technique has overcome all the classical and state-of-the-art descriptors. Although in Brodatz data the margin to improve the classification rate is small, we may notice that the proposed technique provided a more robust result and with a significant advantage over the other methods. Relative to the other statistical measures, we notice that they confirm the correctness efficiency. Particularly, the proposed method also presents a minimum error (1 and 2) and a perfect reliability (until the significance level adopted), both in correct and wrong classifications. We see that, although Bouligand-Minkowski shows similar values, it presents a significantly lower AER value, implying that the classifier confuses descriptors from a relevant number of classes.

Vistex dataset is a more complicated case once it was developed for color analysis approaches. Thus, the use of gray level descriptors is waited to present a defficient result. Nevertheless, even not using color properties, the multi-level descriptors achieved a good classification result, mainly relatively to other gray level texture approaches. Again, as in Brodatz set, the other statistical metrics support the correctness rate. Also, again, the AER in Gabor descriptors is meaningly lower than that of MLD descriptors.
\begin{table*}[!htpb]
	\centering
	\scriptsize
		\begin{tabular}{cccccccc}
			\hline
                 Method & ND & CR (\%) & $\kappa$ & ACR & AER & AE1 & AE2\\
                 \hline
                 \hline
                  Gabor & 20 & 88.19 & 0.88 & 0.96 & 0.79 & 0.11 & 0.12\\
                  Co-occurrence & 24 & 79.63 & 0.79 & 0.95 & 0.79 & 0.17 & 0.20\\
                  GLDM & 20 & 67.36 & 0.67 & 0.84 & 0.67 & 0.31 & 0.33\\
                  Laws & 15 & 84.03 & 0.84 & 0.87 & 0.63 & 0.15 & 0.16\\
                  Multifractal & 101 & 32.41 & 0.32 & 0.89 & 0.82 & 0.63 & 0.68\\
                  Bouligand-Minkowski & 85 & 86.81 & 0.87 & 0.98 & 0.91 & 0.12 & 0.13\\
                  Proposed method & 101 & \underline{92.82} & 0.93 & 0.99 & 0.94 & 0.06 & 0.07\\
                  \hline			
		\end{tabular}
	\caption{Correctness rate for Vistex dataset. The best result is underlined.}
	\label{tab:vistex}
\end{table*}

Generally speaking, we observe in both datasets that the proposed descriptors have presented a performance even better than Bouligand-Minkowski approach. Bouligand-Minkowski method has already demonstrated to be an efficient tool for the discrimination of natural textures. Such performance is explained by the dilation process in that method. As the surface points are dilated, some wavefronts start to emerge. The distribution of these wavefronts provide a rich description of the original arrangement of pixel intensities in the image. Moreover, it still gives information about physical characteristics, like luminance, roughness and even material composition. On the other hand, here we proposed an improvement to the original Bouligand-Minkowski approach. The present technique shows three main additions to the original method. The first is the multiscale decomposition. In this way, we are able to capture localized spatial details with a greater accuracy. The second is the use of deviation descriptors. Actually, Bouligand-Minkowski descriptors are in some sense similar in their general aspect. The deviation allows a more highlighted representation of patterns embedded in the original descriptors and, as a consequence, increases the power of the classifier technique. A third point is the use of entropy measure. Shannon entropy is a classical method to measure the information content in a data. In this way, here, the entropy plays the role of attenuating possible redundancies and describe the original texture with the minimum necessary number of descriptors.

\begin{figure}[!htb] % Figuras lado a lado
					 \centering
           \mbox{\subfigure[]{\epsfig{figure=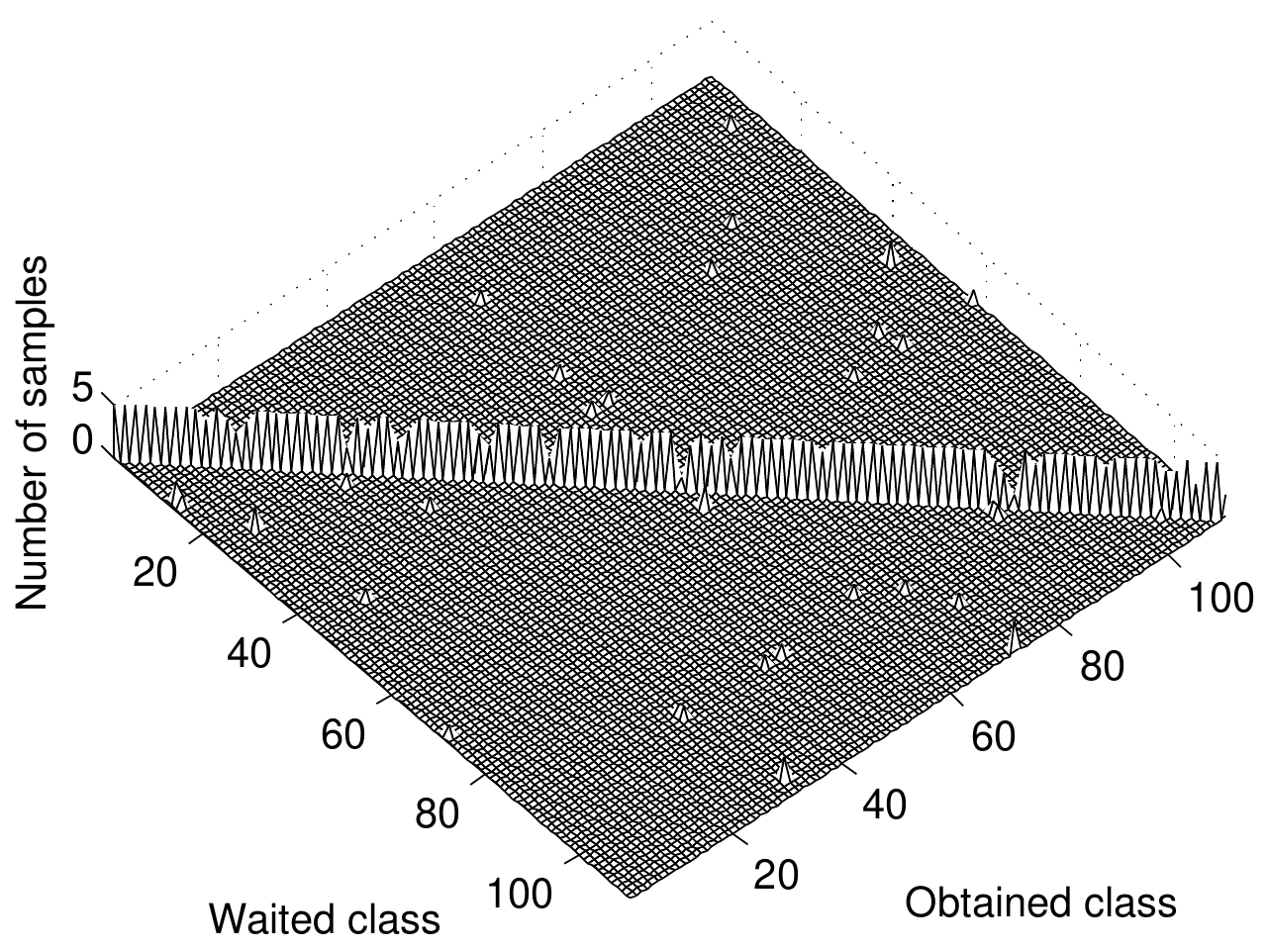,width=0.5\columnwidth}}
                 \subfigure[]{\epsfig{figure=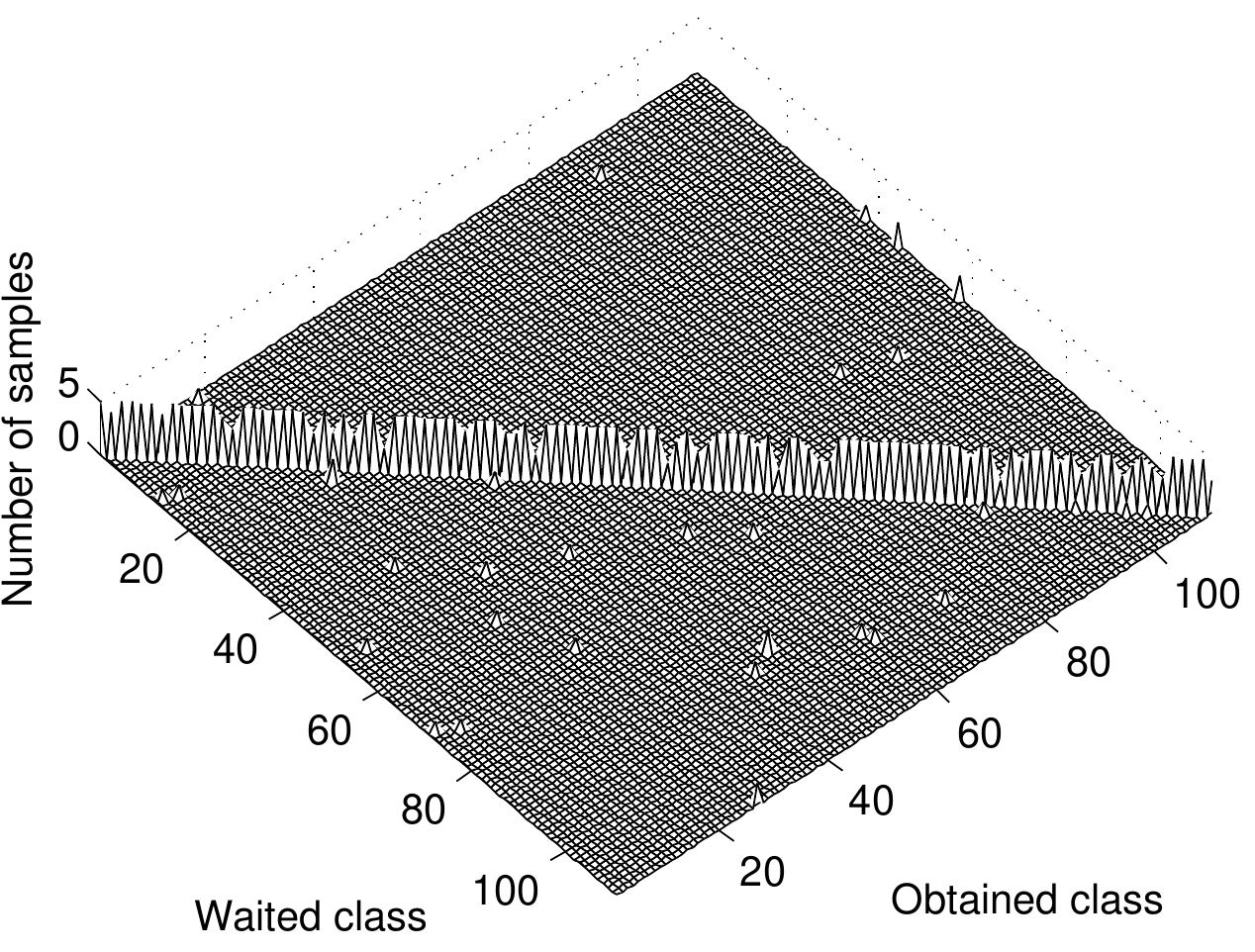,width=0.5\columnwidth}}}
           \mbox{\subfigure[]{\epsfig{figure=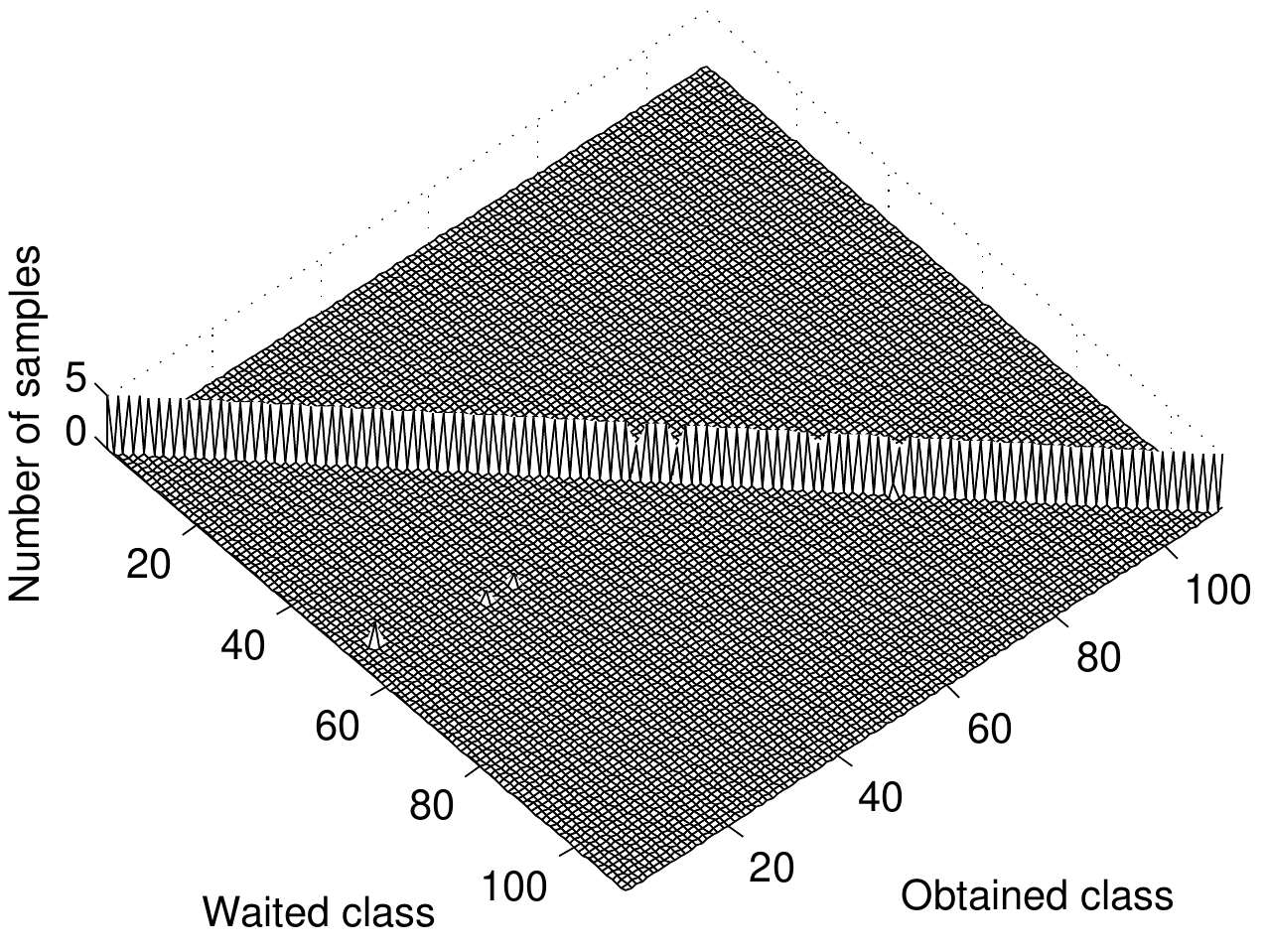,width=0.5\columnwidth}}
           			 \subfigure[]{\epsfig{figure=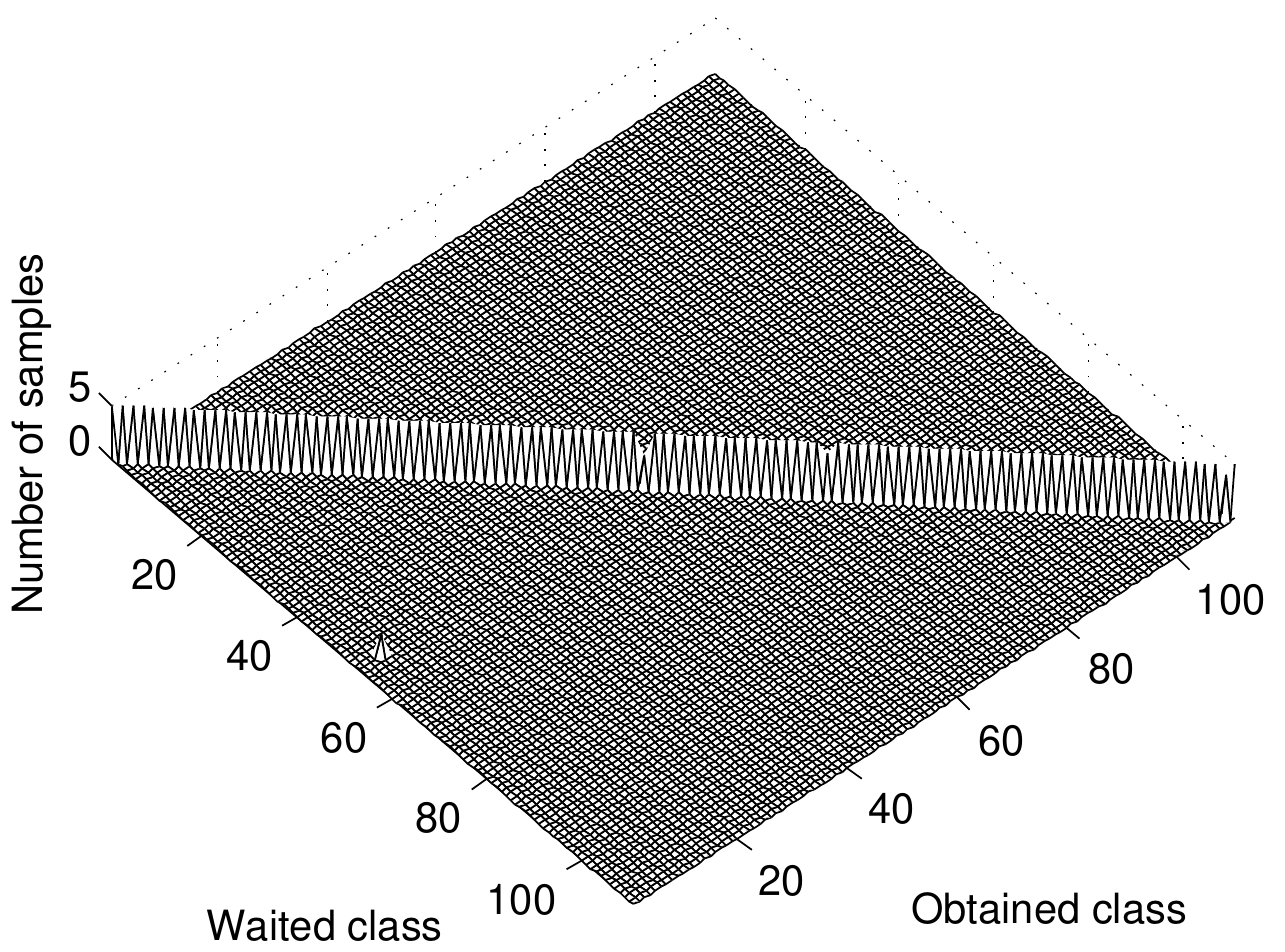,width=0.5\columnwidth}}}           			 
           \caption{Surface representation of confusion matrices in Brodatz dataset. a) Co-occurrence. b) Gabor. c) Bouligand-Minkowski. d) Proposed method. }
           \label{fig:CMbrodatz}                                  
   \end{figure}

\begin{figure}[!htb] % Figuras lado a lado
					 \centering
           \mbox{\subfigure[]{\epsfig{figure=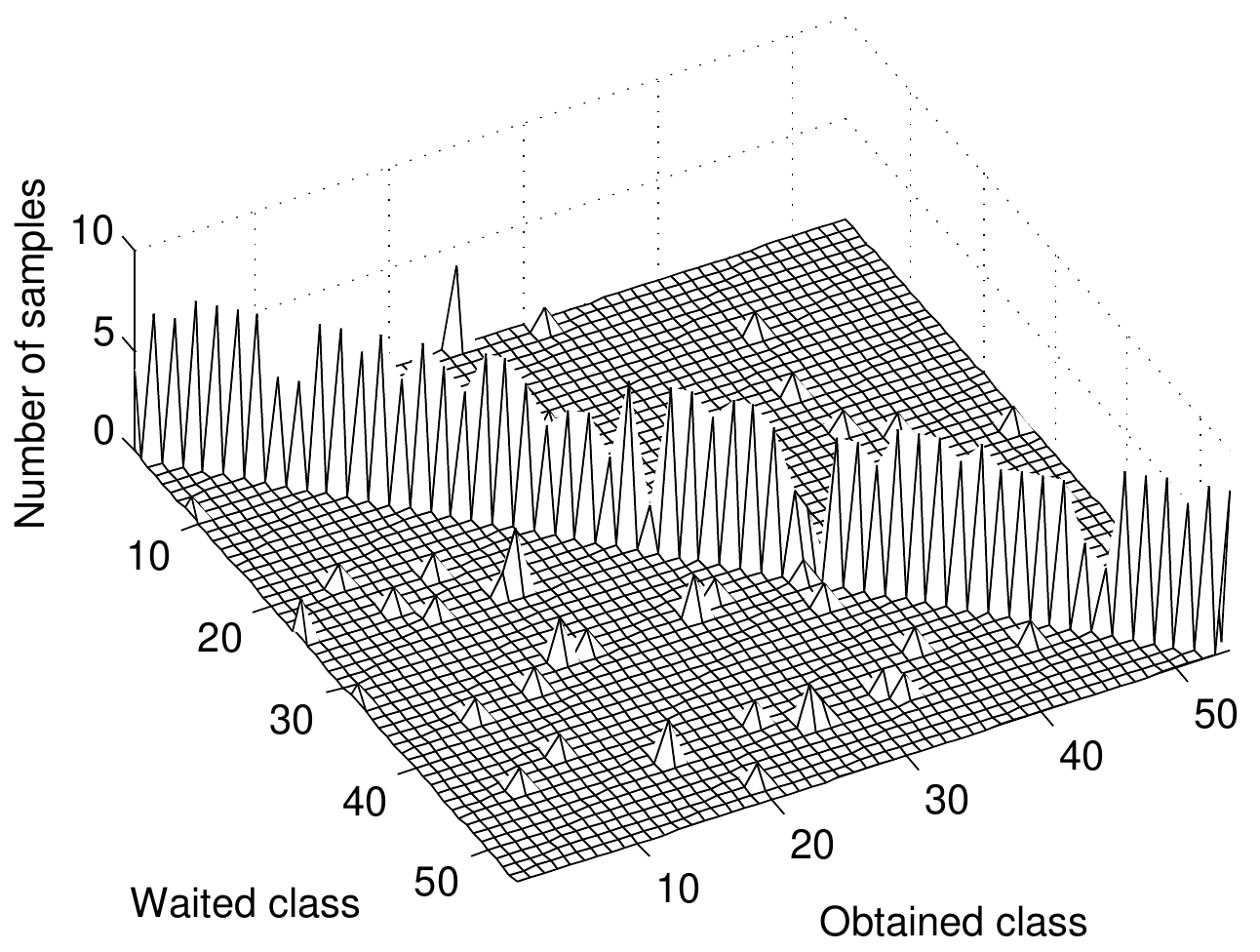,width=0.5\columnwidth}}
                 \subfigure[]{\epsfig{figure=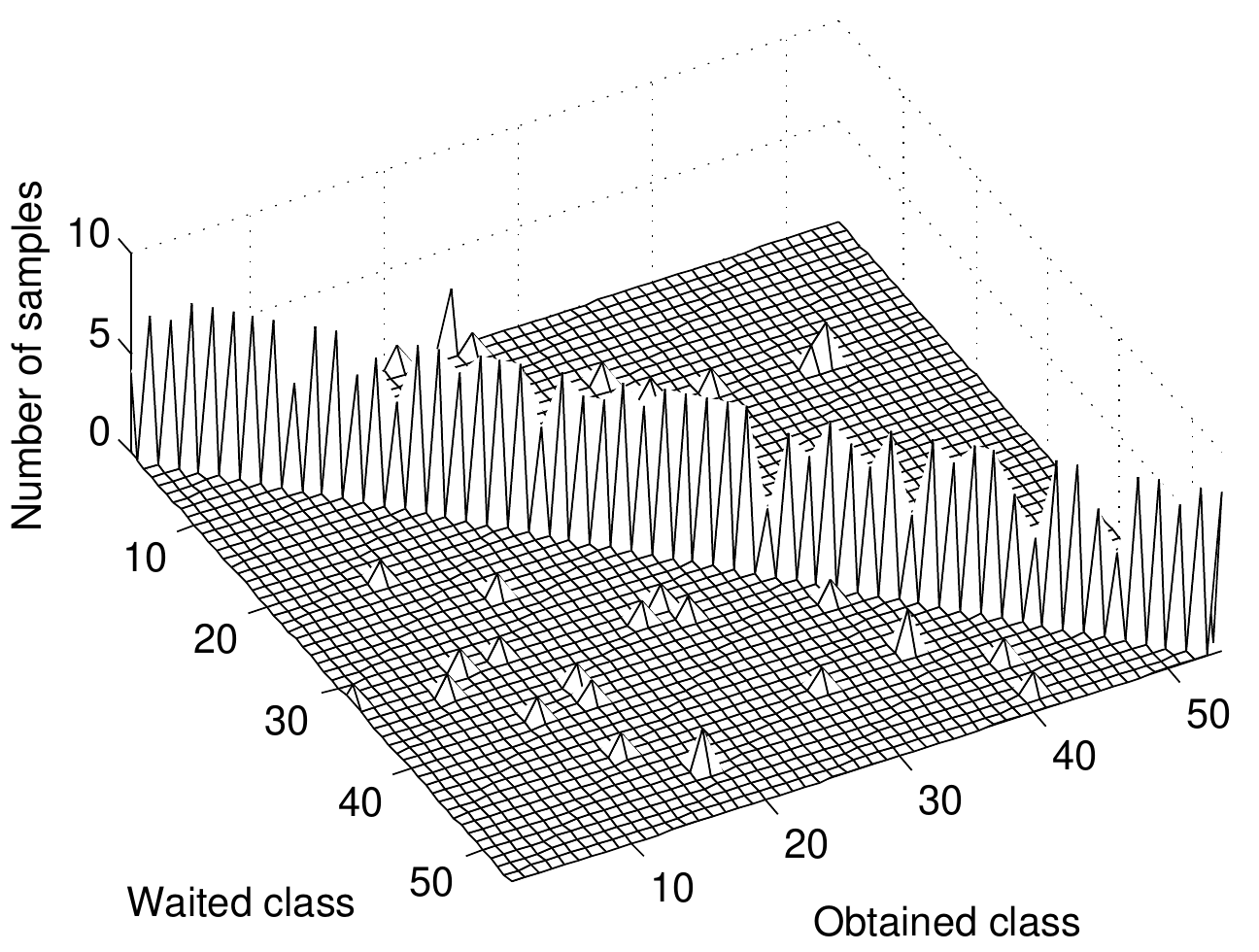,width=0.5\columnwidth}}}
           \mbox{\subfigure[]{\epsfig{figure=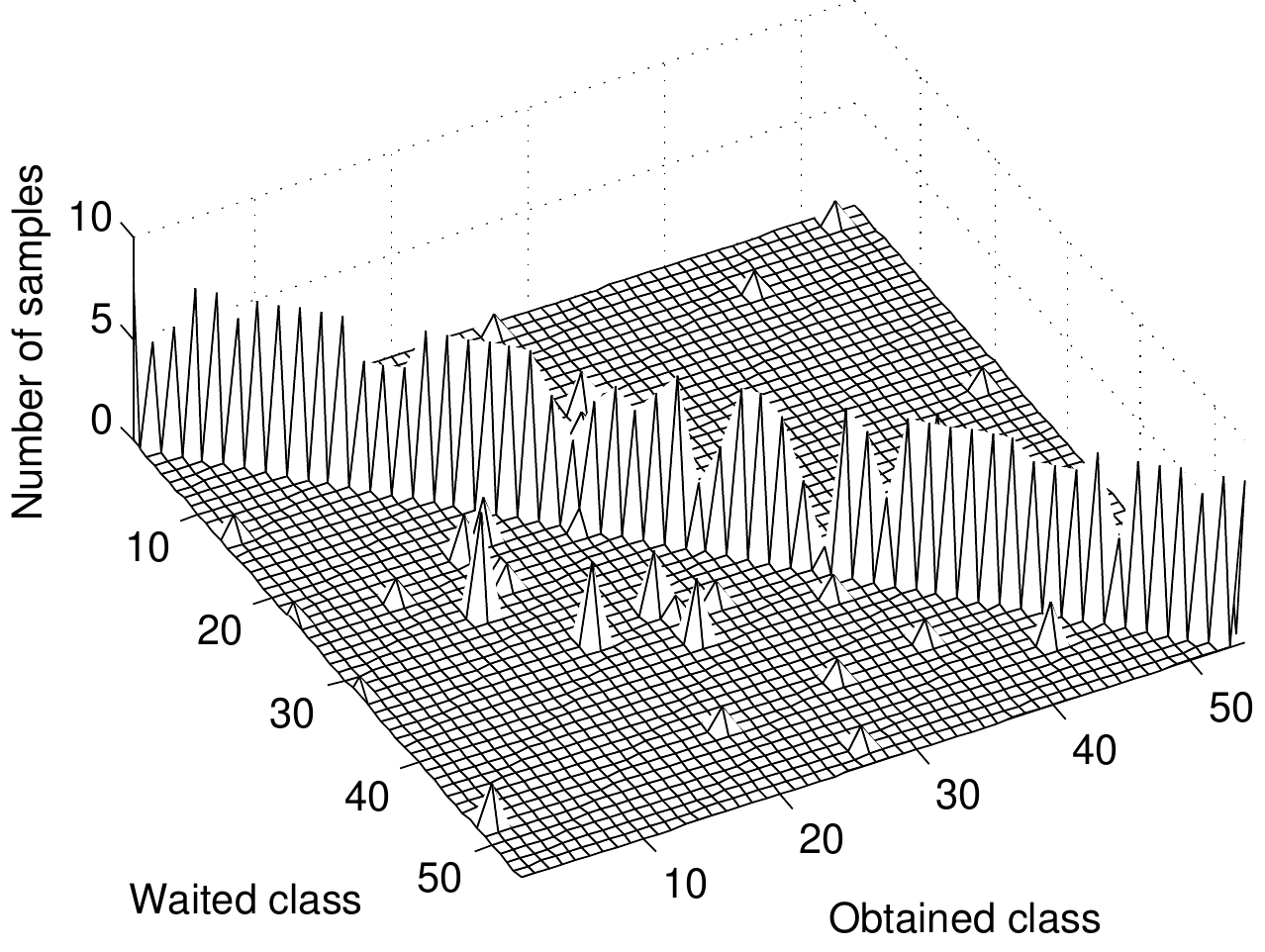,width=0.5\columnwidth}}
           			 \subfigure[]{\epsfig{figure=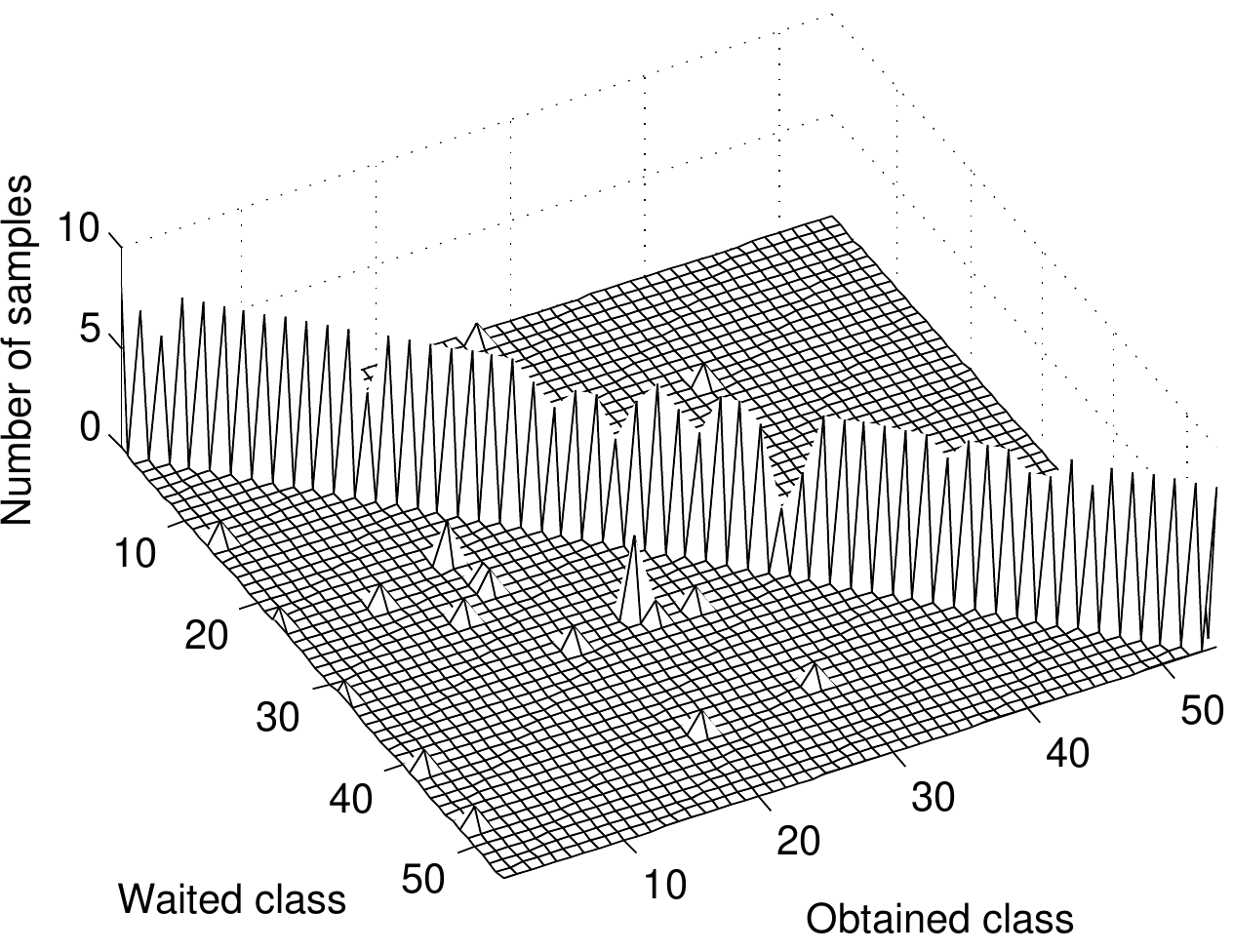,width=0.5\columnwidth}}}           			 
           \caption{Surface representation of confusion matrices in Brodatz dataset. a) Laws. b) Gabor. c) Bouligand-Minkowski. d) Proposed method. }
           \label{fig:CMvistex}                                  
   \end{figure}

Finally, the Figures \ref{fig:CMbrodatz} and \ref{fig:CMvistex} exhibit the confusion matrices for the 4 methods which presented the greater correctness rates in each dataset. In both figures, the matrix is represented through a surface in which the heights correspond to the number of samples classified to class $i$ and pertaining to class $j$. In these surfaces, a good method is represented through a diagonal with a wall aspect (with a minimum amount of ``holes'') and with a minimum number of peaks outside the diagonal. In this sense, for Brodatz dataset, we observe that the MLD matrix shows only one protuberant peak in the lower half of the matrix. In the case of Vistex, we also see a reduced number of peaks outside the diagonal.

\section{Conclusions}

This work develops and study a new approach for texture descriptors based on fractal geometry, more specifically, on Bouligand-Minkowski fractal descriptors. The technique computes the Bouligand-Minkowski descriptors of an image under different decomposition levels and for each level we estimate the average and deviation descriptors. Thus, we extract statistical measures for each average and deviation. These measures compose the feature vector of the texture.

The method was tested on a classification task of benchmark texture datasets and compared to other classical texture descriptors approaches. The results demonstrated the higher accuracy of multi-level descriptors in this task. These results also illustrates the importance of a decomposition step in the application of Bouligand-Minkowski descriptors and the relevance of the deviation descriptors in the discrimination of more complex texture datasets. 

Finally, this outcome suggests that the present technique is a powerful approach to describe and discriminate gray-level textures. This also points to the possibility of using the proposed method in a large number of problems involving texture analysis and related issues.

\section{Acknowledgements}
\label{sec:Acknowledgements}
Odemir M. Bruno gratefully acknowledges the financial support of CNPq (National Council for Scientific and Technological Development, Brazil) (Grant \#308449/2010-0 and \#473893/2010-0) and FAPESP (The State of S\~ao Paulo Research Foundation) (Grant \# 2011/01523-1). Jo\~ao B. Florindo is grateful to CNPq(National Council for Scientific and Technological Development, Brazil) for his doctorate grant.

%\bibliography{Wavelike}

\end{document}